\documentclass{article}
\pdfoutput=1
\usepackage{ijcai19}

\usepackage{times}
\usepackage{soul}
\usepackage{url}
\usepackage[hidelinks]{hyperref}
\usepackage[utf8]{inputenc}
\usepackage[small]{caption}
\usepackage{graphicx}
\usepackage{amsmath}
\usepackage{booktabs}
\usepackage{algorithm}
\usepackage{algorithmic}
\usepackage{stfloats}

\title{ Microsoft AI Challenge India 2018: Learning to Rank Passages for Web Question Answering with Deep Attention Networks\footnote{This work was presented at 2nd Workshop on Humanizing AI (HAI) at IJCAI'19 in Macao, China.}}

\author{
    Chaitanya Sai Alaparthi
    \affiliations
    IIIT-Hyderabad, India \emails
    chaitanyasai.alaparthi@research.iiit.ac.in
}

\begin{document}

\maketitle

\begin{abstract}
  This paper describes our system for The Microsoft AI Challenge India 2018: Ranking Passages for Web Question Answering. The system uses the bi-LSTM network with co-attention mechanism between query and passage representations. Additionally, we use self attention on embeddings to increase the lexical coverage by allowing the system to take union over different embeddings. We also incorporate hand-crafted features to improve the system performance. Our system achieved a Mean Reciprocal Rank (MRR) of 0.67 on eval-1 dataset.

\end{abstract}

\section{Introduction}

Automated Question Answering (QA) is an attractive variation of search where the QA system automatically returns a passage which is an answer to a user's question, instead of giving several links. Ranking the passages is an important step in Web QA systems, where the candidate passages are identified and scored as likely to contain an answer.

To explore the various practical approaches for this problem, Microsoft India organized the evaluation of ranking of passages for a given user question. We participated in Microsoft AI Challenge India 2018\footnote{All practical information, data download links and the results on eval1 dataset can be consulted via the CodaLab website: \url{https://competitions.codalab.org/competitions/20616}} and have secured a position among the top 20 teams. For a given query and a passage pair, our system begins with assigning a score for each passage and normalizes the scores to form a probability distribution of having an answer across the passages in this pair. This will be done for all pairs. The probability distribution of a pair containing $passage_{i}$ and $passage_{j}$ will be stored in $i^{th}$ row and $j^{th}$ column of Probability Distribution Matrix (PDM) represented as $R_{10 \times 10}$. This matrix is then used to compute the ranking of passages using greedy approach.

The rest of the paper is organized as follows: In \autoref{sec:preprocesssteps}, we analyze the data and describe the pre-processing steps. The details of the model are presented in \autoref{sec:mdes}. In \autoref{sec:rankingmech} we describe the document ranking mechanisms we used during inference. Experiments and results are presented in \autoref{sec:exptres}. We conclude this paper in \autoref{sec:conclu}. For the rest of the paper, we use the term document and passage interchangeably.

\section{Data \& Pre-processing steps}\label{sec:preprocesssteps}
The data sets we used were all provided by the competition organizer team, with no other external corpus. The statistics of the given data set are as follows: in total there are 524K samples, where each sample is comprised of a query, 10 documents and a label denoting the suitable document among the 10 documents. We split the data set into training set and dev set containing 519K samples and 5K samples respectively. The total number of words in training set is 2.1M. In all our experiments, we only considered those words whose frequency is at least three. This reduced the vocabulary size to 567K words. The words which are not in our vocabulary were treated as out-of-vocabulary words. Our models were evaluated on eval-1 and eval-2 data sets provided during the competition.

\subsection{Text pre-processing} We applied the following pre-processing steps: The query and document text were tokenized using NLTK word tokenizer \cite{Loper:2002:NNL:1118108.1118117}. We did not perform stemming/lemmatization. Stopwords, all punctuations were removed, and all letters were converted to lowercase.

\subsection{Pre-trained word embedding} We used 3 types of word embeddings: Word2Vec \cite{NIPS2013_5021}, GloVe \cite{pennington2014glove} and FastText \cite{bojanowski2016enriching} in our experiments. We trained all these word embedding models on a corpus obtained from combining all the queries and documents from the training set. For some of our experiments, we also used pre-trained ELMo embeddings\footnote{\url{https://allennlp.org/elmo}} \cite{Peters:2018} but discontinued later due to huge increase in training time.

\subsection{Hand-crafted features} Apart from semantic and lexical features that will be captured from data by the network, we added the following hand-crafted features to improve the model performance:
\begin{itemize}
  \item Sentence length of documents.
  \item TF-IDF, BM25 scores of documents for a given query.
\end{itemize}

\section{Model Description}\label{sec:mdes}
\begin{figure*}[hbt!]
   \centering
     \includegraphics[width=\textwidth]{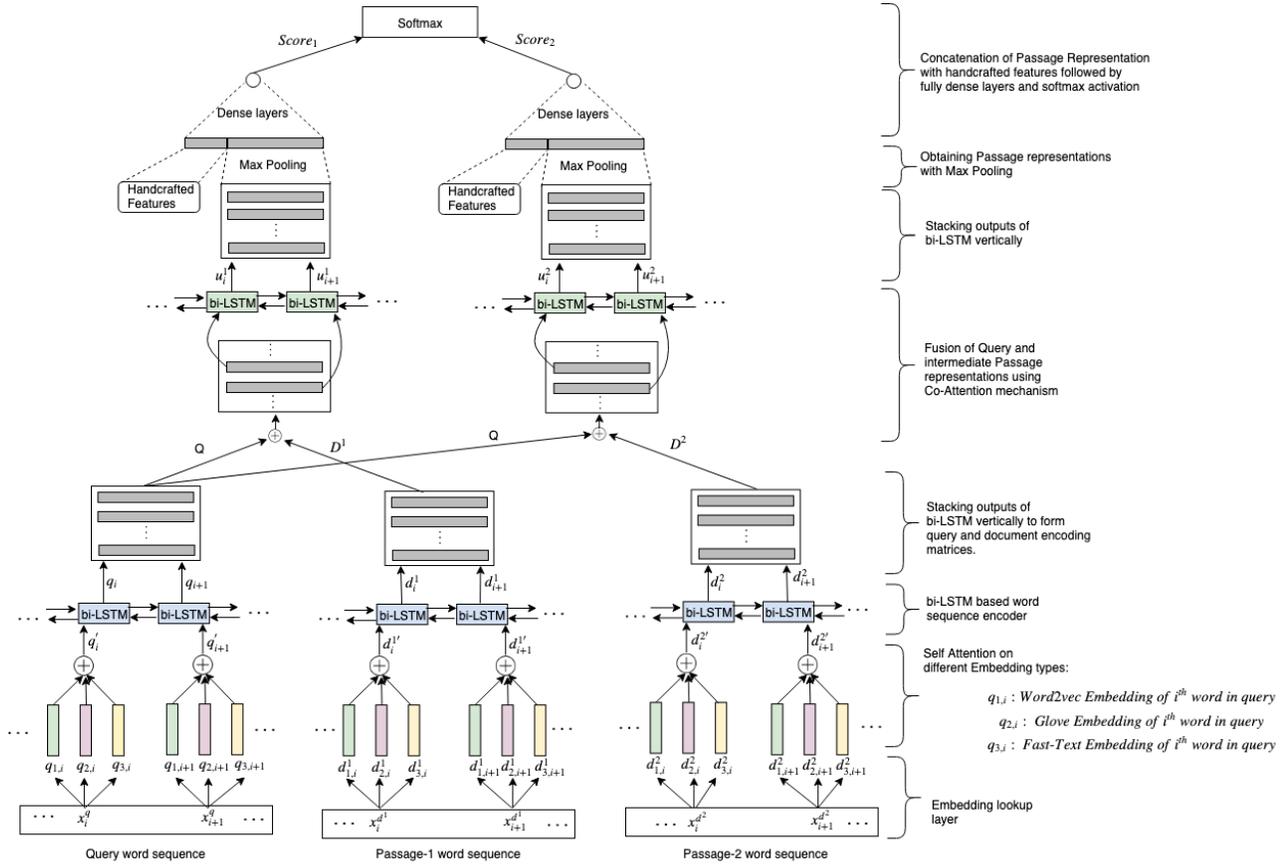}
   \caption{Architecture of best performing model}
   \label{fig:figureA}
\end{figure*}

In this section, we describe the architecture of our best model, shown in \autoref{fig:figureA}. The other architectures we tried are presented in \autoref{sec:modelvariants}. Our system is comprised of 4 parts: (1) The Embedding Layer, where for each word, we look up the Word2Vec, GloVe, FastText embeddings and apply self-attention on these embeddings to get a meta embedding, (2) The bi-LSTM layer, (3) The co-attention layer, where we fuse the intermediate representations of query and a document which were obtained from bi-LSTM layer to obtain query aware document representation, (4) The output layer, where we finally compute the scores and probability distribution of the documents.

\paragraph{Notations.} For each query, we take a single pair of documents and we pass the data to the network as a tuple containing $\langle q$, $d^{1}$, $d^{2} \rangle$. Where $q$ is the query, $d^{1}$ and $d^{2}$ are the possible documents for this query. One of $d^{1}, d^{2}$ is actually the document containing the answer, denoted as $d^{+}$, while the other is not, denoted as $d^{-}$. For each query, we randomly select $d^{-}$ from the set of $\{d^{-}_{j}; j = 1, \ldots , 9\}$. And we randomly shuffle the order between $d^{+}$ and $d^{-}$ to be $d^{1}$ and $d^{2}$ to prevent network from over fitting.

Let $(x^{q}_{1}, x^{q}_{2}, \ldots , x^{q}_{n})$ denote the sequence of words in the $query$ and $(x^{d^{1}}_{1}, x^{d^{1}}_{2}, \ldots , x^{d^{1}}_{m})$, $(x^{d^{2}}_{1}, x^{d^{2}}_{2}, \ldots , x^{d^{2}}_{k})$ denote the same for $document_1$ and $document_2$. Note that $n$, $m$, $k$ are lengths of $query$, $document_{1}$ and $document_{2}$ respectively.

\subsection{Embedding Layer}\label{emblayer}
In this layer, for each word $x^{q}_{i}$ in query $q$, we performed an embedding lookup to get the Word2Vec, GloVe, FastText embeddings represented by $q_{i, 1}$, $q_{i, 2}$, $q_{i, 3}$ respectively. These embeddings were fixed and were pre-trained on the corpus obtained from combining all the queries and documents from the training set. Instead of concatenating these 3 embeddings into a single embedding, inspired from \cite{1508.04257}, \cite{kiela2018dynamic}, we combined these embeddings by taking the weighted average denoted as:
\begin{equation}
q^{'}_{i} = \sum_{j=1}^{3} \alpha_{i, j}q_{i, j}
\end{equation}

where $\alpha_{i, j}$ are scalar weights from self-attention mechanism: 
\begin{equation}
a_{i, j} = \textbf{W}.q_{i, j} + \textbf{b}
\end{equation}
\begin{equation}
\alpha_{i, j} = \frac{exp(a_{i, j})}{\sum_{k=1}^{3}exp(a_{i, k})}
\end{equation}
Here $q^{'}_{i}$ is the meta-embedding of $i^{th}$ word in query which can be fed into biLSTM. Here $\textbf{W}$ and $\textbf{b}$ are learnable parameters. Similarly, we perform the embedding lookup and self-attention on the words from documents $d_{1}$ and $d_{2}$ using the same set of parameters $\textbf{W}$ and $\textbf{b}$ to obtain $d^{1'}$ and $d^{2'}$.

The main reason we used weighted average instead of concatenation is because: enhancement of performance and improved coverage of vocabulary \cite{1508.04257}.

\subsection{Bi-LSTM Layer}
We used a standard bidirectional LSTM \cite{Hochreiter:1997:LSM:1246443.1246450} encoder to encode the query and document sequence of word vectors as: 
\begin{equation}
q_{i} = biLSTM_{enc}(q_{i-1}, q_{i+1}, q^{'}_{i})
\end{equation}
\begin{equation}
d^{1}_{i} = biLSTM_{enc}(d^{1}_{i-1}, d^{1}_{i+1}, d^{1'}_{i})
\end{equation}
\begin{equation}
d^{2}_{i} = biLSTM_{enc}(d^{2}_{i-1}, d^{2}_{i+1}, d^{2'}_{i})
\end{equation}

The query and document sequences are computed with the same $biLSTM_{enc}$ to share the representation power \cite{Mueller:2016:SRA:3016100.3016291}. From the outputs of $biLSTM_{enc}$, we define the query encoding matrix as $Q=[q_{1}, q_{2}, \ldots , q_{n}, q_{\phi}]$. We also added sentinel vector  $q_{\phi}$ \cite{1609.07843} which allows the model to not attend to any particular word in the input. Similarly, we define document encoding matrices as
$D^{1}=[d^{1}_{1}, d^{1}_{2}, \ldots , d^{1}_{m}, d_{\phi}]$ and $D^{2}=[d^{2}_{1}, d^{2}_{2}, \ldots , d^{2}_{k}, d_{\phi}]$.
\subsection{Co-Attention Layer}
Inspired from \cite{1606.00061} and \cite{1611.01604}, we used the same co-attention mechanism that attends to the query and document simultaneously and finally fuses both attention contexts. We first describe co-attention in general and apply it to
$\langle Q, D^{1}\rangle$ and $\langle Q, D^{2}\rangle$ to get query aware document representations $U^{1}$ and $U^{2}$.

Similar to \cite{1611.01604}, we first compute the affinity matrix, which contains affinity scores corresponding to all pairs of query and document words: $L = D^{T}Q$. The affinity matrix is normalized row wise to get the attention weights $A^{Q}$ across the document for each word in query. And similarly normalized column-wise to get the attention weights $A^{D}$ across the query for each word in the document:
\begin{equation}
A^{Q} = softmax(L)
\end{equation}
and 
\begin{equation}
A^{D} = softmax(L^{T})
\end{equation}

Next, we compute the attention contexts, of the document in light of each word of the query:
\begin{equation}
C^{Q} = DA^{Q}
\end{equation}
We similarly compute the attention contexts $QA^{D}$ of the query in light of each word of the document. Additionally, we compute the summaries $C^{Q}A^{D}$ of the previous attention contexts in light of each word of the document. We also define $C^{D}$, a co-dependent representation of the query and document, as the co-attention context:
\begin{equation}
C^{D} = [Q; C^{Q}]A^{D}
\end{equation}
where $[x;y]$ is concatenation of vectors $x$ and $y$ horizontally.
The last step is the fusion of temporal information to the co-attention context via a bidirectional LSTM:
\begin{equation}
  u_{i} = biLSTM_{fusion}(u_{i-1}, u_{i+1}, [d_{i};c^{D}_{i}])  
\end{equation}
We define $U=[u_{1}, \ldots , u_{l}]$, the outputs of $biLSTM_{fusion}$ concatenated vertically, which provides a foundation for finding the likelihood of the document containing the answer, as the co-attention encoding.

Using this mechanism on $\langle Q, D^{1}\rangle$ and $\langle Q, D^{2}\rangle$, we obtain
\begin{equation}
  U^{1} = Co\text{-}Attention(Q, D^{1})  
\end{equation}
and 
\begin{equation}
  U^{2} = Co\text{-}Attention(Q, D^{2})  
\end{equation}

\subsection{Output Layer}
In this layer, we first apply the max-pooling operation on $U^{1}$ and $U^{2}$ to get the final representations of documents, followed by concatenation of manual features: document length, BM25, TF-IDF scores:
\begin{equation}
  V_{1} = max(\{u^{1}_{i}\}_{i=1, 2, \ldots , m})  
\end{equation}
\begin{equation}
  V_{1} = [length ; BM25 ; tf\text{-}idf ; V_{1}]  
\end{equation}
We then obtain a linear transformation on $V_{1}$ to get a score for this $document_{1}$:
\begin{equation}
  score_{1} = \textbf{W}_{s}.{V_{1}} + \textbf{b}_{s}  
\end{equation}

Similarly, we apply the same for $document_{2}$:
\begin{equation}
    V_{2} = max(\{u^{2}_{i}\}_{i=1, 2, \ldots , k})
\end{equation}
\begin{equation}
  V_{2} = [length ; BM25 ; tf\text{-}idf ; V_{2}]  
\end{equation}
\begin{equation}
  score_{2} = \textbf{W}_{s}.{V_{2}} + \textbf{b}_{s}  
\end{equation}

We used these scores $score_{1}$ and $score_{2}$ to compute the posterior probability of a document containing an answer given a query through a softmax function:
\begin{equation}
  P(D^{i}|\langle Q, D^{1}, D^{2} \rangle) = \frac{exp(score_{i})}{\sum_{j=1}^{2}exp(score_{j})}  
\end{equation}

During training, the model parameters were estimated to maximize the likelihood of the document which has answer in it, given the queries across the training set. Equivalently, we need to minimize the following loss function:
\begin{equation}
  L(\Theta) = -\sum_{\langle Q, D^{+}, D^{-} \rangle \: \in \: \textbf{S}} log P(D^{+}|\langle Q, D^{+}, D^{-} \rangle))  
\end{equation}
Where \textbf{S} is the training set, $\Theta$ denotes the parameters set of the neural network. Since $L(\Theta)$ is differentiable with respect to $\Theta$, the model is trained readily using gradient-based numerical optimization algorithms.
\section{Ranking Mechanisms}\label{sec:rankingmech}
During inference, for a given query and 10 corresponding documents, we compute the Probability Distribution Matrix (PDM) denoted as $R_{10 \times 10}$, where $i^{th}$ row and $j^{th}$ column of the matrix is given by $R_{i, j}$:
\begin{equation}
  R_{i, j} = P(D^{i}|\langle Q, D^{i}, D^{j} \rangle)  
\end{equation}
One can observe that:
\begin{equation}
  R_{j, i} = P(D^{j}|\langle Q, D^{i}, D^{j} \rangle) = 1 - P(D^{i}|\langle Q, D^{i}, D^{j} \rangle)  
\end{equation}

The reason behind using PDM is that, our model could only rank between two documents at once. So to predict the ranks between all the ten documents, we first compute the probability distribution (ranks) among the documents in each pair and store them in the PDM. We then apply heuristics to rank all the ten documents at once.

Let us suppose $R_{i, j} > R_{j, i}$, it means that $document_{i}$ is better than $document_{j}$ in containing answer, implying rank of $document_{i}$ is better than rank of $document_{j}$. Ideally, this corresponds to finding the ranks of documents which strictly satisfies every pair in PDM. But there could be a possibility that no ranking sequence exists strictly satisfying every pair in PDM. Because the PDM does not guarantee to satisfy transitive property\footnote{Suppose $R_{i, j} > R_{j, i}$ and $R_{j, k} > R_{k, j}$, but it \textbf{need not be true} that $R_{i, k} > R_{k, i}$ as the values of PDM comes from a neural network which does not guarantee the transitivity.}. To overcome that, we can relax the strict condition and find the ranks of documents which satisfies most number of pairs in PDM. This problem can be thought of as finding Hamiltonian path with maximum path sum on a complete digraph $K_{10}$, where the edge weight from $vertex_{i}$ to $vertex_{j}$ is 1 if $R_{i, j} > R_{j, i}$ else 0. This problem is $\mathcal {NP} $-hard and the time complexity of a solution involving dynamic programming is $O(N^{2}.2^{N})$. $N$ being the number of vertices in the graph, which in our case is 10. As the number of samples are huge in $eval_1$ and $eval_2$ data sets, we decided not to use this approach as the overall time taken to calculate the ranks could be very high. Instead, we used the sub-optimal greedy approach described in \autoref{alg:algorithm1}, which runs in $O(N^2)$.

\begin{algorithm}[tb]
\caption{Greedy Ranking}\label{alg:algorithm1}
\textbf{Input}: $R_{10 \times 10}$\\
\textbf{Output}: $rank$, denoting ranks of 10 documents. \\
\begin{algorithmic}[1] 
\STATE {$A_{i}$ $\gets$ {$0$} $\forall i \in [1, 10]$}
\STATE {$rank_{i}$ $\gets$ {$0$} $\forall i \in [1, 10]$}
\FOR{$i \gets 1$ to $10$} 
\STATE    $A_{i} \gets \sum_{j=1}^{10}R_{i, j}$
\ENDFOR
\STATE $order \gets \underset{(max)}{argsort(A)}$
\FOR{$i \gets 1$ to $10$} 
\STATE    $rank_{order_{i}} \gets i$
\ENDFOR
\STATE \textbf{return} $rank$
\end{algorithmic}
\end{algorithm}

\section{Experiments and Results}\label{sec:exptres}
In this section, we report our evaluation results based on the given eval-1\footnote{\url{https://competitions.codalab.org/competitions/20616}} data set.
\subsection{Model variants}\label{sec:modelvariants}
We conduct experiments with the following variants:
\begin{itemize}
    \item  \textbf{M1} Bi-LSTM sentence encoder with GloVe embeddings and without co-attention, which is our basic system.
    \item \textbf{M2} Bi-LSTM sentence encoder with ELMo embeddings \cite{Peters:2018} and without co-attention.
    \item \textbf{M3} Same as \textit{M2} but applied co-attention between query and document.
    \item \textbf{M4} Bi-LSTM sentence encoder with Word2Vec, FastText and GloVe embeddings. All these embeddings were concatenated. In addition to these, we used manual features such as sentence lengths of documents, TF-IDF, BM25 scores of document for a given query. We applied co-attention between query and document.
    \item \textbf{M5} Our best model described in \autoref{sec:mdes}.
\end{itemize}
\subsection{Implementation Details}
We trained and evaluated our models on the data set provided by the competition organizer team, with no other external corpus. The dimension of word embeddings was set to 300 in all our experiments and the embeddings were fixed during the training. We limited the vocabulary to the words whose frequency is at least three and set embeddings of Word2Vec and GloVe for out-of-vocabulary words to zero.

We set the maximum query length as 15 and maximum document length as 70. The hidden state size was set to 500 and number of layers were fixed to 2 in all our experiments for all bi-LSTMs. All our network parameters were randomly initialized uniformly in the range $[-0.01, 0.01]$. We used dropout of 0.2 between the LSTM layers to regularize our network during training \cite{JMLR:v15:srivastava14a}, and optimized the model using ADAM \cite{kingma:adam}, with initial learning rate of $0.001$. We normalized the L2-norm of the gradient of the cost function each time to be at most a predefined threshold of 5, when the norm was larger than the threshold \cite{Pascanu:2013:DTR:3042817.3043083}. We trained for 100 epochs with each batch of size 256. All our models were implemented and trained using PyTorch \cite{paszke2017automatic}.

\begin{table}
\centering
\begin{tabular}{lll}
\hline
Model  & MRR \\
\hline
BM25 Baseline & 0.43     \\
DL Baseline & 0.48      \\
M1 & 0.49      \\
M2 & 0.55      \\
M3 & 0.58      \\
M4 & 0.63      \\
M5 & \textbf{0.67}    \\
\hline
\end{tabular}
\caption{MRR scores of different models we used on eval-1 dataset.}
\label{tab:plain}
\end{table}

\subsection{Results}
\autoref{tab:plain} shows the results on eval-1 dataset. Along with our models, We include two versions of the baselines provided during the competition: BM25 Baseline and DL Baseline.

We experimented with different model variants that we introduced in \autoref{sec:modelvariants} to analyze the effectiveness of each model. We first compare \textit{M1}, \textit{M2}. The reason \textit{M2} performed much better than \textit{M1} while the only major difference being the word embeddings is because, ELMo representations are purely character based, allowing the network to use morphological clues to form robust representations for out-of-vocabulary tokens unseen in training. In case of \textit{M1}, the embedding of out-of-vocabulary words are zeros leading the network to under fit.

Next we compare models \textit{M2} and \textit{M3}. The only difference between them is use of co-attention. From the results, we can conclude that co-attention mechanism is helping the network to perform much better by capturing the relation between a query and a document. Although models with ELMo embeddings are performing better, they are taking huge amount of time to train leading to difficulty of experimenting with other ideas. For this reason, we switched back to traditional embeddings for the next set of experiments. However, instead of experimenting with GloVe embeddings alone, we considered Word2Vec and FastText to overcome the drawback of out-of-vocabulary words and to increase the representation power of the neural network. The use of FastText is important as it can get the representation of an out-of-vocabulary word with the help of sub-word information. With out it, the problem of out-of-vocabulary words would still be there. In addition to these embeddings, we also used few manual features described in \autoref{sec:preprocesssteps}. With these embeddings and manual features, the model \textit{M4} outperformed previous models by a huge margin, making room for new set of experiments. 

We then improved \textit{M4} to obtain \textit{M5} by applying self-attention mechanism on embeddings as described in \autoref{emblayer}. This led to further improvement of score as can be seen in \autoref{tab:plain}.

\section{Conclusion}\label{sec:conclu}
In this paper, we described our models that we used in Microsoft AI Challenge India 2018. We used the bi-LSTM with co-attention mechanism between query and a document. In addition to co-attention, we also used self-attention mechanism on different embeddings types leading to further improvement of our model performance.

In future work, it would be interesting to use context aware embeddings such as ELMo. It would also be interesting to further improve our model by replacing recurrent models with transformer networks \cite{devlin2018bert}. Additionally, it would be interesting to explore other useful hand-crafted features and ensembling methods.

\bibliographystyle{named}
\bibliography{ijcai19}

\begin{thebibliography}{}

\bibitem[\protect\citeauthoryear{Bojanowski \bgroup \em et al.\egroup
  }{2016}]{bojanowski2016enriching}
Piotr Bojanowski, Edouard Grave, Armand Joulin, and Tomas Mikolov.
\newblock Enriching word vectors with subword information.
\newblock {\em arXiv preprint arXiv:1607.04606}, 2016.

\bibitem[\protect\citeauthoryear{Devlin \bgroup \em et al.\egroup
  }{2018}]{devlin2018bert}
Jacob Devlin, Ming-Wei Chang, Kenton Lee, and Kristina Toutanova.
\newblock Bert: Pre-training of deep bidirectional transformers for language
  understanding.
\newblock {\em arXiv preprint arXiv:1810.04805}, 2018.

\bibitem[\protect\citeauthoryear{Hochreiter and
  Schmidhuber}{1997}]{Hochreiter:1997:LSM:1246443.1246450}
Sepp Hochreiter and J\"{u}rgen Schmidhuber.
\newblock Long short-term memory.
\newblock {\em Neural Comput.}, 9(8):1735--1780, November 1997.

\bibitem[\protect\citeauthoryear{Kiela \bgroup \em et al.\egroup
  }{2018}]{kiela2018dynamic}
Douwe Kiela, Changhan Wang, and Kyunghyun Cho.
\newblock Dynamic meta-embeddings for improved sentence representations.
\newblock In {\em Proceedings of the 2018 Conference on Empirical Methods in
  Natural Language Processing (EMNLP)}, Brussels, Belgium, 2018.

\bibitem[\protect\citeauthoryear{Kingma and Ba}{2015}]{kingma:adam}
Diederick~P Kingma and Jimmy Ba.
\newblock Adam: A method for stochastic optimization.
\newblock In {\em International Conference on Learning Representations (ICLR)},
  2015.

\bibitem[\protect\citeauthoryear{Loper and
  Bird}{2002}]{Loper:2002:NNL:1118108.1118117}
Edward Loper and Steven Bird.
\newblock Nltk: The natural language toolkit.
\newblock In {\em Proceedings of the ACL-02 Workshop on Effective Tools and
  Methodologies for Teaching Natural Language Processing and Computational
  Linguistics - Volume 1}, ETMTNLP '02, pages 63--70, Stroudsburg, PA, USA,
  2002. Association for Computational Linguistics.

\bibitem[\protect\citeauthoryear{Lu \bgroup \em et al.\egroup
  }{2016}]{1606.00061}
Jiasen Lu, Jianwei Yang, Dhruv Batra, and Devi Parikh.
\newblock Hierarchical question-image co-attention for visual question
  answering, 2016.

\bibitem[\protect\citeauthoryear{Merity \bgroup \em et al.\egroup
  }{2016}]{1609.07843}
Stephen Merity, Caiming Xiong, James Bradbury, and Richard Socher.
\newblock Pointer sentinel mixture models, 2016.

\bibitem[\protect\citeauthoryear{Mikolov \bgroup \em et al.\egroup
  }{2013}]{NIPS2013_5021}
Tomas Mikolov, Ilya Sutskever, Kai Chen, Greg~S Corrado, and Jeff Dean.
\newblock Distributed representations of words and phrases and their
  compositionality.
\newblock In C.~J.~C. Burges, L.~Bottou, M.~Welling, Z.~Ghahramani, and K.~Q.
  Weinberger, editors, {\em Advances in Neural Information Processing Systems
  26}, pages 3111--3119. Curran Associates, Inc., 2013.

\bibitem[\protect\citeauthoryear{Mueller and
  Thyagarajan}{2016}]{Mueller:2016:SRA:3016100.3016291}
Jonas Mueller and Aditya Thyagarajan.
\newblock Siamese recurrent architectures for learning sentence similarity.
\newblock In {\em Proceedings of the Thirtieth AAAI Conference on Artificial
  Intelligence}, AAAI'16, pages 2786--2792. AAAI Press, 2016.

\bibitem[\protect\citeauthoryear{Pascanu \bgroup \em et al.\egroup
  }{2013}]{Pascanu:2013:DTR:3042817.3043083}
Razvan Pascanu, Tomas Mikolov, and Yoshua Bengio.
\newblock On the difficulty of training recurrent neural networks.
\newblock In {\em Proceedings of the 30th International Conference on
  International Conference on Machine Learning - Volume 28}, ICML'13, pages
  III--1310--III--1318. JMLR.org, 2013.

\bibitem[\protect\citeauthoryear{Paszke \bgroup \em et al.\egroup
  }{2017}]{paszke2017automatic}
Adam Paszke, Sam Gross, Soumith Chintala, Gregory Chanan, Edward Yang, Zachary
  DeVito, Zeming Lin, Alban Desmaison, Luca Antiga, and Adam Lerer.
\newblock Automatic differentiation in pytorch.
\newblock In {\em NIPS-W}, 2017.

\bibitem[\protect\citeauthoryear{Pennington \bgroup \em et al.\egroup
  }{2014}]{pennington2014glove}
Jeffrey Pennington, Richard Socher, and Christopher~D. Manning.
\newblock Glove: Global vectors for word representation.
\newblock In {\em Empirical Methods in Natural Language Processing (EMNLP)},
  pages 1532--1543, 2014.

\bibitem[\protect\citeauthoryear{Peters \bgroup \em et al.\egroup
  }{2018}]{Peters:2018}
Matthew~E. Peters, Mark Neumann, Mohit Iyyer, Matt Gardner, Christopher Clark,
  Kenton Lee, and Luke Zettlemoyer.
\newblock Deep contextualized word representations.
\newblock In {\em Proc. of NAACL}, 2018.

\bibitem[\protect\citeauthoryear{Srivastava \bgroup \em et al.\egroup
  }{2014}]{JMLR:v15:srivastava14a}
Nitish Srivastava, Geoffrey Hinton, Alex Krizhevsky, Ilya Sutskever, and Ruslan
  Salakhutdinov.
\newblock Dropout: A simple way to prevent neural networks from overfitting.
\newblock {\em Journal of Machine Learning Research}, 15:1929--1958, 2014.

\bibitem[\protect\citeauthoryear{Xiong \bgroup \em et al.\egroup
  }{2016}]{1611.01604}
Caiming Xiong, Victor Zhong, and Richard Socher.
\newblock Dynamic coattention networks for question answering, 2016.

\bibitem[\protect\citeauthoryear{Yin and Schütze}{2015}]{1508.04257}
Wenpeng Yin and Hinrich Schütze.
\newblock Learning meta-embeddings by using ensembles of embedding sets, 2015.

\end{thebibliography}
\end{document}